\documentclass{article}

\usepackage[nonatbib, preprint]{neurips_2023}

\usepackage[utf8]{inputenc} 
\usepackage[T1]{fontenc}    
\usepackage{hyperref}       
\usepackage{url}            
\usepackage{booktabs}       
\usepackage{amsfonts}       
\usepackage{nicefrac}       
\usepackage{microtype}      

\usepackage{multirow}
\usepackage[caption=false]{subfig}
\usepackage{graphicx}

\title{Unlocking the Potential of Federated Learning for Deeper Models}

\author{Haolin Wang$^{1}$,~ Xuefeng Liu$^{1,2}$\thanks{For correspondence, please contact: liu\_xuefeng@buaa.edu.cn},~ Jianwei Niu$^{1,2,3}$,~ Shaojie Tang$^{4}$,~ Jiaxing Shen$^{5}$ \\ \\
	$^1$Beihang University, $^2$Zhongguancun Laboratory, $^3$Zhengzhou University \\$^4$University of Texas at Dallas, $^5$Lingnan University
}

\begin{document}
	
	\maketitle
	
	\begin{abstract}
		Federated learning (FL) is a new paradigm for distributed machine learning that allows a global model to be trained across multiple clients without compromising their privacy. Although FL has demonstrated remarkable success in various scenarios, recent studies mainly utilize shallow and small neural networks. In our research, we discover a significant performance decline when applying the existing FL framework to deeper neural networks, even when client data are independently and identically distributed (i.i.d.). Our further investigation shows that the decline is due to the continuous accumulation of dissimilarities among client models during the layer-by-layer back-propagation process, which we refer to as "divergence accumulation." As deeper models involve a longer chain of divergence accumulation, they tend to manifest greater divergence, subsequently leading to performance decline. Both theoretical derivations and empirical evidence are proposed to support the existence of divergence accumulation and its amplified effects in deeper models. To address this issue, we propose several technical guidelines based on reducing divergence, such as using wider models and reducing the receptive field. These approaches can greatly improve the accuracy of FL on deeper models. For example, the application of these guidelines can boost the ResNet101 model's performance by as much as 43\% on the Tiny-ImageNet dataset.
	\end{abstract}
	
	\section{Introduction}
	\label{introduction}
	
	Federated learning (FL) is an emerging distributed learning framework that allows a global model to be trained across multiple clients and without privacy leakage\cite{fedavg, FL1, FL2, FL3}.
	While recent FL studies achieve notable success in various contexts, they primarily utilize shallow, small-scale neural networks with typically less than ten layers \cite{fedavg, small1, small2, small3}. In contrast, centralized learning often enjoys larger and deeper models due to their increased capacity for fitting a diversity of data. For example, ResNet101\cite{resnet} has 101 layers, Swin-L\cite{swin} has 120 layers and DeepNet even has 1000 layers \cite{deepnet}. 
	
\begin{figure*}[!t]
		\centering
		\subfloat[Test results for Tiny-ImageNet.]{\includegraphics[width=1.7in]{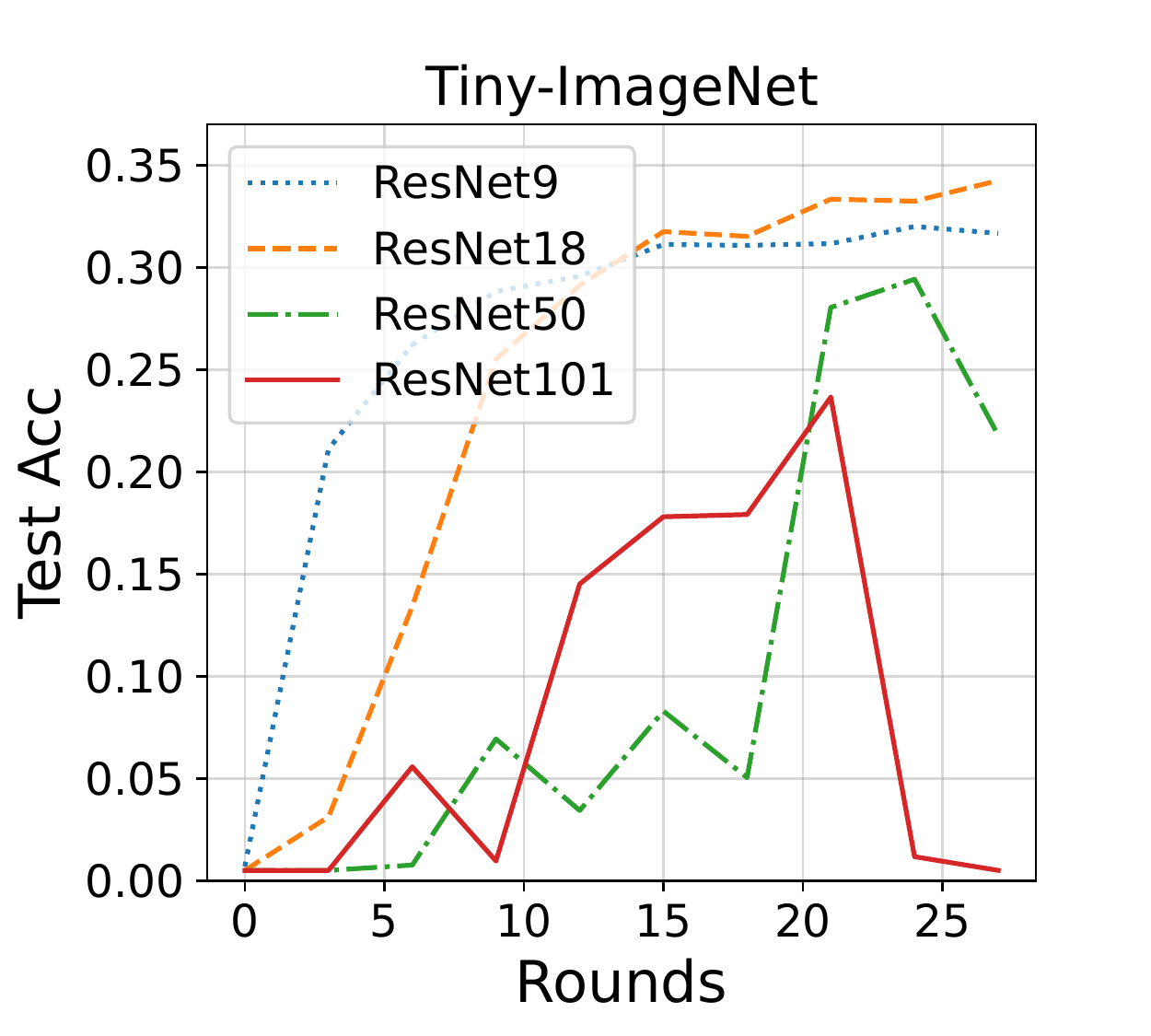}%
			\label{fig:imagenet_tiny}}
		\hfil
		\subfloat[Test results for CIFAR100.]{\includegraphics[width=1.7in]{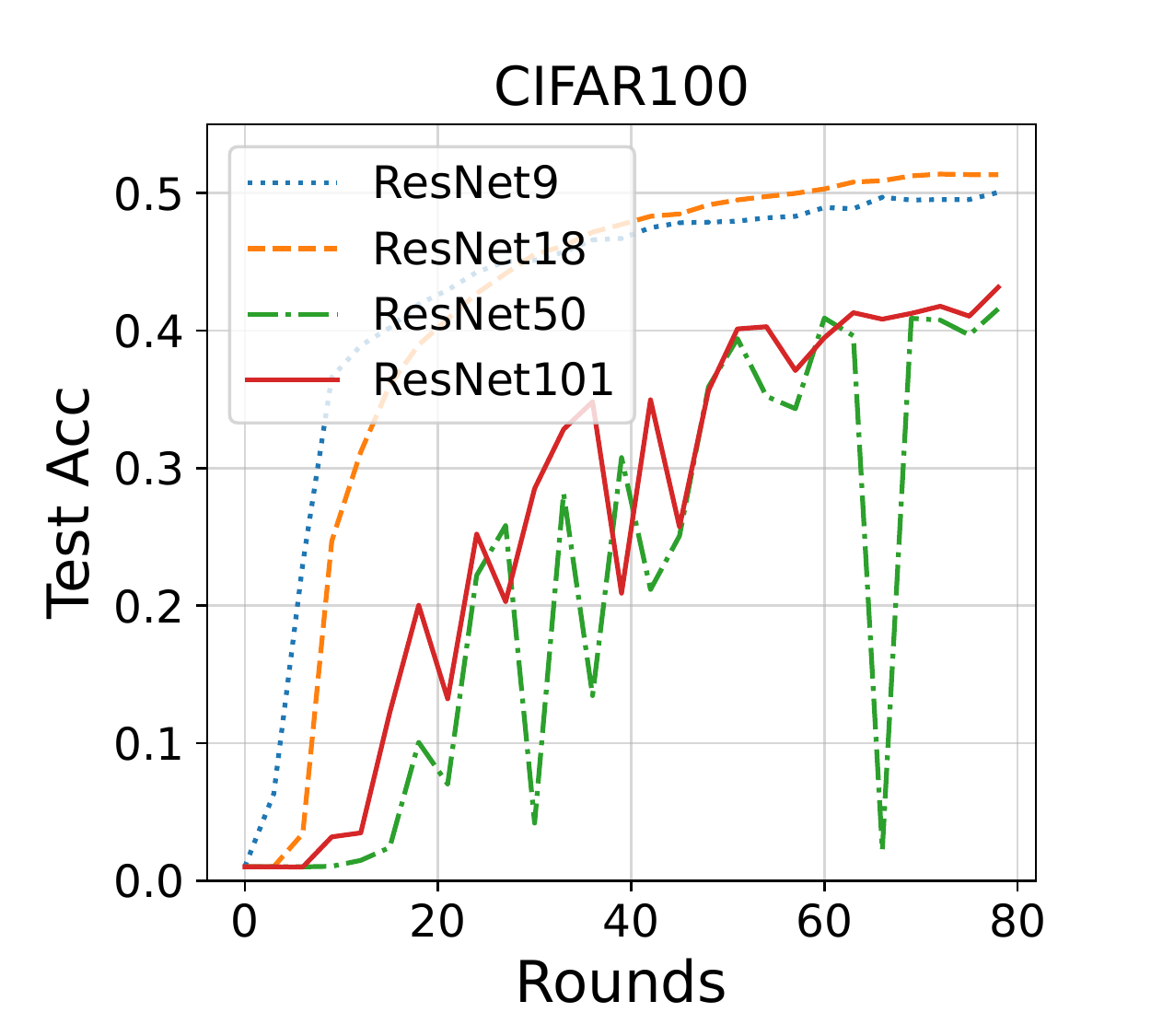}%
			\label{fig:cifar100}}
		\hfil
		\subfloat[Test results for CIFAR10.]{\includegraphics[width=1.7in]{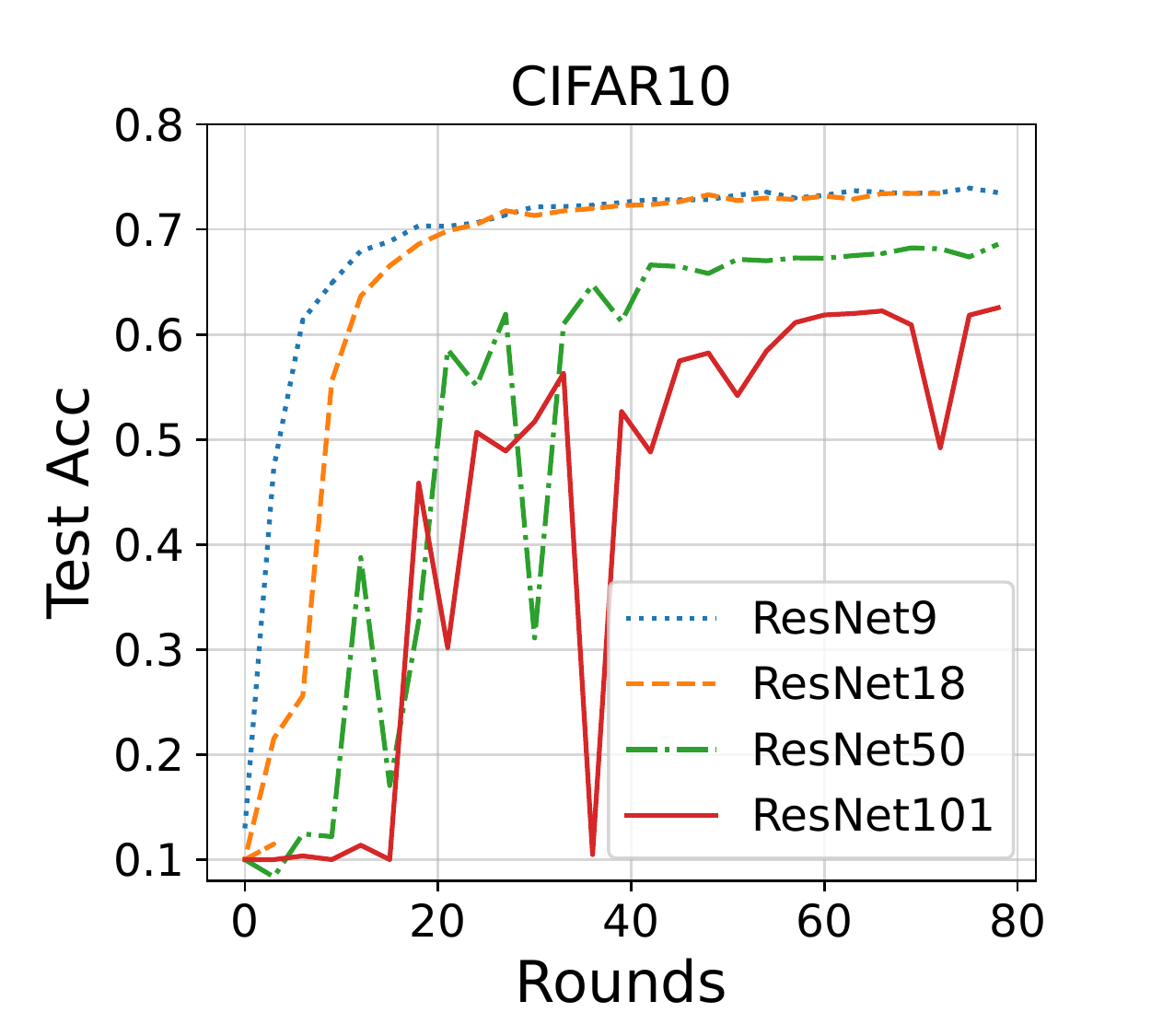}%
			\label{fig:cifar10}}
		\caption{This experiment is conducted on various datasets using ResNets\cite{resnet} with different depths. We choose 30 clients, 8 local epochs and learning rate to be 0.02. The test accuracies on the Tiny-ImageNet dataset using the corresponding models in centralized learning are: 0.383, 0.40, 0.380, 0.340 respectively; the test accuracies on CIFAR100 are: 0.561, 0.552, 0.530, 0.510 respectively; the test accuracies on CIFAR10 are: 0.808, 0.777, 0.752, 0.728 respectively. The performance degradation in FL is noticeably more significant compared to centralized learning.}
		\label{fig:diff_depth}
	\end{figure*}

	Naturally, this architecture gap raises our curiosity about the performance of these deeper architectures within the FL framework. We conduct experiments on models with various depths, and unfortunately, we observe that the performance of FL often deteriorates on a large scale when the neural network becomes deeper, even in a simple context where data across clients are independently and identically distributed (i.i.d.). According to Fig. \ref{fig:diff_depth}, there is an obvious decline for network accuracy (16-33\%), convergence speed (33-89\%) and the training stability. This phenomenon impedes the adoption of FL in many applications where deeper models are usually required \cite{drive_app, medical_app, sur_app}. 
	
	One straightforward approach to enhance FL performance involves making the FL process more similar to centralized learning by reducing the number of local iterations or decreasing the learning rate. This is based on the fact that when the number of local iterations in FL is set to one, it becomes equivalent to centralized learning. However, these methods contradict the core motivation of FL and result in more frequent communications. This not only impacts the efficiency of FL but also raises concerns about privacy protection\cite{privacy}.
	
	To address this problem and achieve stable training with deeper models in FL, we need to answer a crucial question: \emph{Why does using deeper networks in FL lead to a more significant performance decline compared with using shallower networks?}
	
	One conjecture suggests that deeper models with a larger number of parameters are more prone to overfitting the local data on each client, resulting in a degradation of overall performance. However, our experiments cast doubt on this conjecture. If it were true, increasing the number of model parameters, either by widening or deepening the model, should consistently lead to a decline in performance. Surprisingly, when we increase the number of model parameters by widening each layer, we observe a large improvement in FL performance instead of a decline. This indicates that attributing the performance degradation solely to the number of parameters is insufficient. For more detailed information on this experiment, refer to Section \ref{section:experiment}.
	
	To provide a clearer understanding of the aforementioned question, we introduce the concept of divergence for client models in FL. Divergence, in this context, signifies the dissimilarity or distance between different clients and serves as a significant indicator for evaluating FL performance. We prove that if the global optimal point coincides with the local optimal point of each client, the optimization objective of FL will align with that of centralized learning. A detailed proof can be found in Section \ref{theorem}. Thus, a lower divergence implies a closer approximation of FL model to centralized model, which further infers enhanced performance of this FL model. Formally, we define divergence as follows. For training round $t$ and model layer $L$, the average divergence of $N$ clients is:
	\begin{equation}
		div_t^L=\frac 1 N \sum_{i=1}^N \sqrt{\frac{||w_{i,t}^L - \bar w_t^L||_2^2}{d^L}}, \quad \bar w_t^L=\frac 1 N \sum_{i=1}^{N} w_{i,t}^L \quad \forall w_{i,t}^L\in\mathbb{R}^{d^L} \label{eq:div1}
	\end{equation}
	where $w_{i,t}^L$ is the parameter of the $L$th layer in client $i$ at round $t$, $\bar w_t^L$ is the corresponding global model, $d^L$ is the dimension of $w_{i,t}^L$. 
	
\begin{figure*}[!t]
		\centering
		\subfloat[Normalized divergences in shallow layers.]{\includegraphics[width=2.75in]{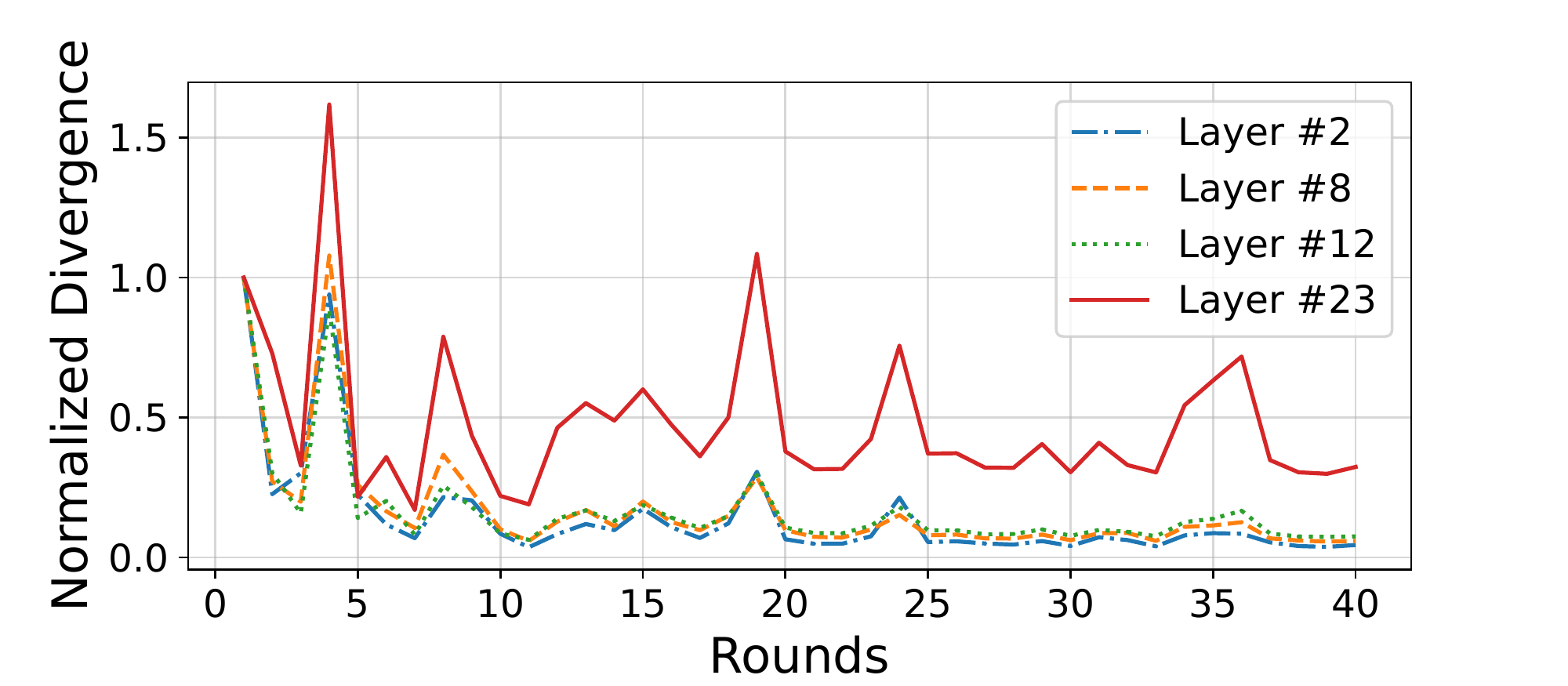}%
			\label{fig:shallow_std}}
		\hfil
		\subfloat[Normalized divergences in deep layers.]{\includegraphics[width=2.75in]{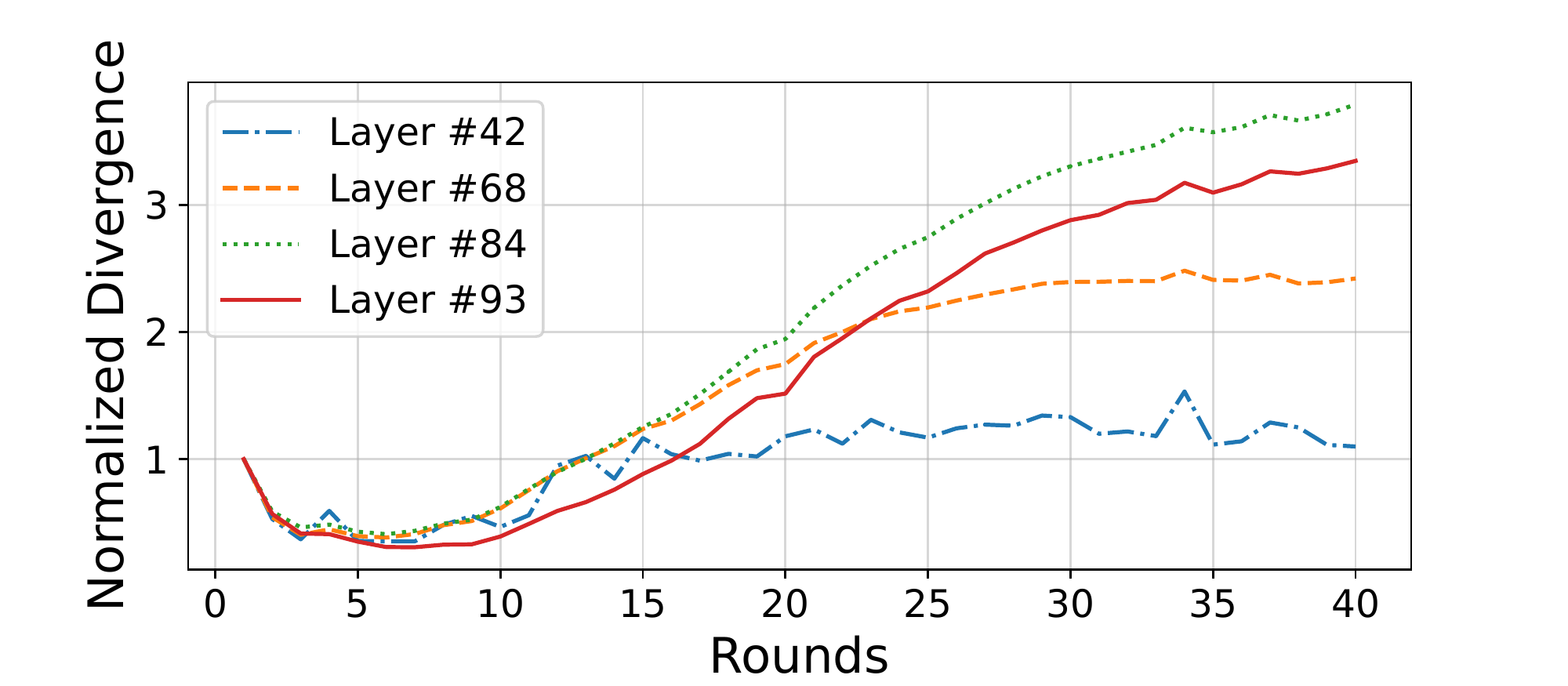}%
			\label{fig:deep_std}}
		\caption{In this experiment, we demonstrate the divergences of different layers in ResNet101. To visualize the evolving pattern of divergence for each layer, we normalize each $div_t^L$ by dividing $div_1^L$. This ensures that each line starts from the value of 1. (a) shows that the divergence of shallow layers gradually decline, while (b) shows that divergence of deep layers tends to intensify. }
		\label{fig:diff_std}
	\end{figure*}

	We demonstrate the divergences of ResNet101 trained on Tiny-ImageNet with client number to be 30, and the results are shown in Fig. \ref{fig:diff_std}. Here are several noteworthy observations derived from this figure. Firstly, the divergences of various layers exhibit distinct properties. In the case of shallow layers, divergences generally decrease and eventually converge. However, for deep layers, the divergences tend to increase and intensify. Secondly, it is evident that the deeper layers do not reach convergence prior to the shallower layers.
	
\begin{figure}[t]
		\centering
		\includegraphics[width = 0.8\textwidth]{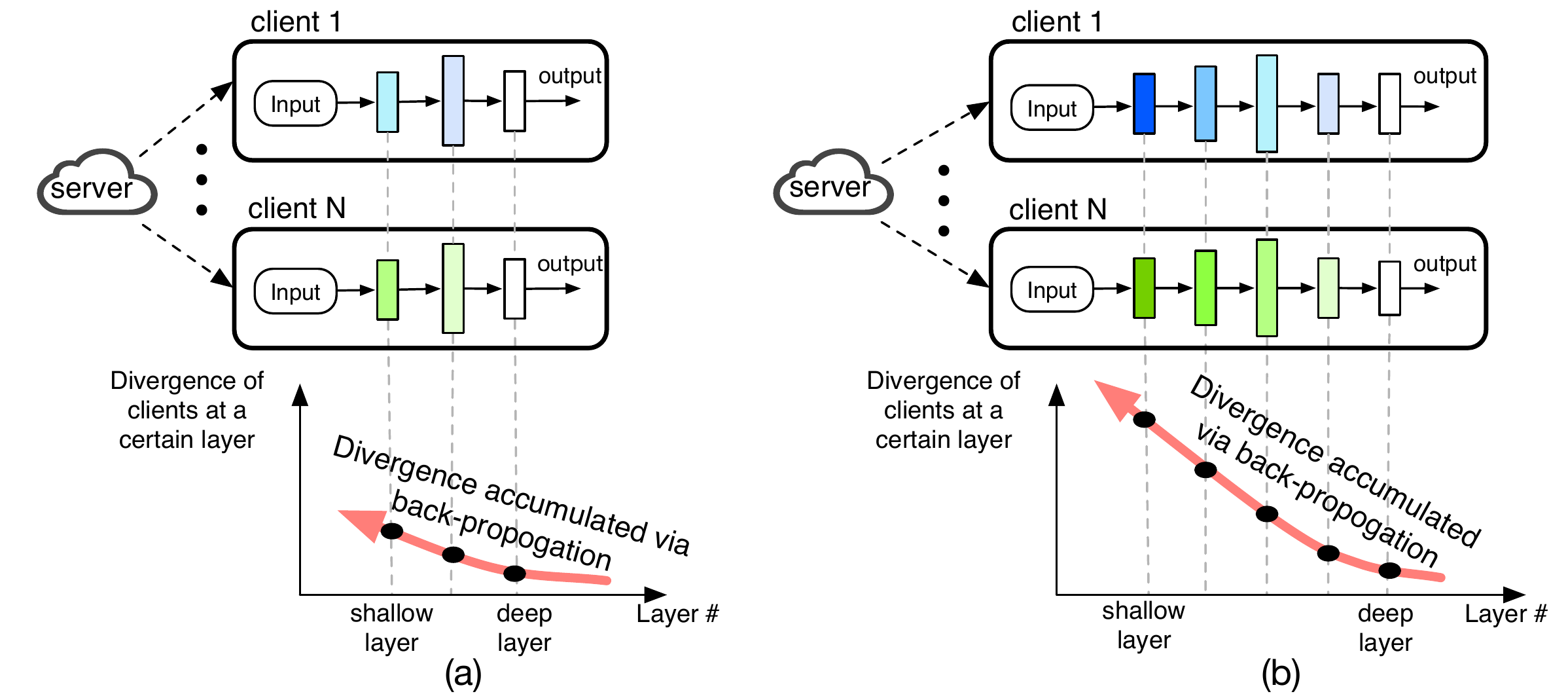}
		\caption{The process of divergence accumulation in (a) a shallow neural network and in (b) a deeper network. As the latter has a longer accumulation chain, it tends to present a larger divergence, especially at shallower layers. This explains the low performance of deeper models in FL.}
		\label{fig:accumulation}
	\end{figure}

	These findings emphasize a strong relationship between parameter divergence and model architecture. To shed light on this phenomenon, we introduce a theorem centered around divergence accumulation. The fundamental idea is illustrated in Fig. \ref{fig:accumulation}. In essence, the divergence of each layer is influenced by the back-propagation algorithm and can be dissected into two components: \emph{the divergence back-propagated from the subsequent layer} and \emph{the divergence originating from distinct inputs}. As each layer integrates divergence from the subsequent layer, divergences accumulate at each step of the back-propagation process. Through this layer-by-layer progression, the divergences of later layers steadily accumulate upon the divergences of preceding layers. Consequently, \emph{deeper networks exhibit longer chains of divergence accumulation}, ultimately resulting in lower performance.
	
	Building upon the aforementioned theorem, we propose several guidelines to enhance the training of deeper networks in the FL framework, focusing on two key aspects. The first aspect relates to \emph{model architecture}. We have discovered that using wider models and reducing receptive fields can mitigate the adverse effects of divergence accumulation and therefore significantly improve FL performance. The second aspect concerns the \emph{pre-processing of input data}. As the fundamental cause of model divergence is the diversity of data originating from different clients, implementing appropriate pre-processing strategies, including using images with higher resolution and adopting proper data augmentation methods, can mitigate the divergence at the root level, thus substantially enhancing the performance of deeper models in FL.
	
	It is crucial to emphasize that our focus is not on developing new methods for enhancing model performance. Instead, we draw attention to existing techniques that are already utilized in centralized learning. However, these methods often come with higher computational costs and offer only marginal improvements in centralized learning scenarios. Nevertheless, our experiments reveal that when applied in the context of FL, these methods have a remarkable impact in reducing divergence, resulting in significantly greater enhancements in FL performance compared to centralized learning. By highlighting these guidelines, our objective is to enhance the understanding of the key considerations in FL as compared to centralized learning. In summary, the contributions of this paper are the following:
	
	First, we observe a noteworthy phenomenon: deeper models in the context of FL often face challenges in achieving convergence, resulting in a degradation of performance. We consider this issue to be critical, especially in the implementation of large-scale FL systems. To the best of our knowledge, our paper is the first to systematically investigate this phenomenon, exploring its underlying causes and proposing potential solutions.
	
	Second, we introduce the concept of "divergence" as a metric to quantify the dissimilarity in update directions across clients in FL. Utilizing this measurement, we make an important observation: divergences for shallow layers generally decrease and eventually converge, while divergences for deep layers tend to increase and intensify. This finding sheds light on the behavior and performance characteristics of different layers in FL models.	
	
	Third, to offer a rationale for the aforementioned observation, we present a theorem that specifically addresses the notion of divergence accumulation. To gain a thorough understanding of the process of divergence accumulation in the context of FL, we provide both experimental evidence and theoretical analysis.
	
	Finally, we put forth several principles that serve as guidelines for reducing divergence and enhancing the performance of FL, particularly in the context of deeper models. By adhering to these principles, practitioners can effectively improve the performance and convergence of FL systems when working with deeper models.
	
	\section{Related Work}
	\textbf{Federated Learning with deeper neural networks.} There are a few works applying deeper models to FL. For example, FetchSGD \cite{fetchsgd} uses ResNet101 to train on FEMNIST \cite{femnist}. Since its main purpose is to reduce communication costs, they did not give an analysis of performance degradation using deeper models. Other works utilize pretrained large models\cite{fedcv, disentangled, ensemble} and their goal is using FL to finetune on various datasets. This setting is different from ours that trains a global model collaboratively from scratch. Some other works \cite{covid19, sparsefed} train deeper models with high-resolution images (e.g. CT images with resolution of $1024\times1024$). Our analysis shows that image with high resolution is beneficial for deeper models in FL. In summary, although there are currently some works using deeper models and achieving considerable results in different FL scenarios, they either require some specific conditions (such as using pretrained models) or need to satisfy certain properties (such as high-resolution images). 
	To the best of our knowledge, there is no previous work that systematically analyzes the causes and solutions for the difficulties in applying deeper models to FL. 
	
	\textbf{Neural Architecture Search (NAS) for Federated Learning.} 
	The goal of NAS is to find the optimal model architecture for a specific task. In the field of FL, NAS needs to consider not only model performance, but also additional factors, such as the computational capabilities of edge devices and communication overhead. Many works have addressed these challenges and achieve good results\cite{fednas1, fednas2, fednas3, fednas4}. However, due to the large search space of deep networks, most of the research focuses on shallow network design, leading to a lack of studies on  deeper network structures.
	
	\section{Phenomenon of Divergence Accumulation}
	\label{theorem}
	\subsection{Divergence}
	To begin introducing our theory, we first clarify the concept of divergence. In essence, the intuition of divergence is to measure the distinctions in optimization objectives between FL and centralized learning. For centralized training, the optimal model parameters $\theta^*$ satisfies:
	\begin{equation}
		\mathbb{E}_{x\in X}[\nabla_{\theta^*} L(\theta^*, x)] = 0,
	\end{equation}
	where $X$ is the dataset and $L(\cdot)$ is the loss function. In FL, on the other hand, the optimal global model parameter $\theta^*$ satisfies:
	\begin{equation}
		\sum_{i=1}^N\mathbb{E}_{x\in X_i}[\nabla_{\theta^*} L(\theta^*, x)] =0,
	\end{equation}
	where $X_i$ is dataset of the i'th client. Assume datasets across clients are i.i.d, we can derive that the optimal solution of FL equals to that of centralized learning when and only when:
	\begin{equation}
		\mathbb{E}_{x\in X_i}[\nabla_{\theta^*} L(\theta^*, x)] =0, \forall i\in 1,\cdots, N
	\end{equation}
	In other words, if the global minimum point aligns exactly with the local minimum point of each client, the optimization target for FL will be identical to that of centralized learning. This deduction gives us an important insight: if there is a large divergence between the various client models, that is, the global minimum point is far from the local minimum point, then FL may lead to a significant performance degradation compared with centralized learning, and vice versa. 
	This is our intuition for proposing the concept of divergence, which is formally defined in Eq. \ref{eq:div1}.
	
	\subsection{Phenomenon of Divergence Accumulation}
	In this section, we introduce a theorem based on the previously defined divergence, which provides insights into the causes of the performance decline observed in deeper models. Our objective is to establish a connection between the diversity of data and the divergence of model. Formally, we aim to prove that the expected divergence is accumulated during the back-propagation process, that is, the expectation of divergence in a shallow layer is always greater than that in a deep layer.
	
	To begin our analysis, we examine the data. Considering that data across different clients can be seen as random variables, we assume their distributions are identical. For instance, in a classification task, we can represent the data in client $i$ belonging to class $c$ as follows:
	\begin{equation}
		\mathbf{X_i^c}=\mathbf{\bar X^c} + \mathbf{\tilde X_i^c},\quad \mathbf{\tilde X_i^c}\sim p(\mathbf{\tilde X}|c)
	\end{equation}
	
	In the given equation, $\mathbf{\tilde X_i^c}$ represents a random variable that is associated with class $c$ and reflects the diversity of data across different clients. On the other hand, $\mathbf{\bar X^c}$ is also associated with class $c$ but represents the generality or typicality of that class. It can be understood as a prototype concept similar to what has been introduced in previous literature \cite{proto1, proto2, proto3}.
	
	We next consider the model. We use an example involving two linear layers. The forward calculation process can be described as follows:
	\begin{equation}
		\mathbf{H_{i-1}}=\mathbf{A_{i-1}}\cdot\mathbf{Z_{i-2}} + \mathbf{b_{i-1}},\quad
		\mathbf{Z_{i-1}}=\sigma(\mathbf{H_{i-1}}) ,\quad
		\mathbf{H_{i}}=\mathbf{A_{i}}\cdot\mathbf{Z_{i-1}} + \mathbf{b_{i}}
	\end{equation}
	Assume the loss function is $L(\cdot)$, according to the chain rule of derivation, we have:
	\begin{eqnarray}
		\frac{\partial L}{\partial \mathbf{A_{i}}}&=&\frac{\partial L}{\partial \mathbf{H_{i}}}\frac{\partial \mathbf{H_{i}}}{\partial \mathbf{A_{i}}}=\frac{\partial L}{\partial \mathbf{H_{i}}} {\mathbf{Z_{i-1}}^T} \label{eq:7}\\
		\frac{\partial L}{\partial \mathbf{Z_{i-1}}}&=&\frac{\partial L}{\partial \mathbf{H_{i}}}\frac{\partial \mathbf{H_{i}}}{\partial \mathbf{Z_{i-1}}}={\mathbf{A_{i}^T}}\frac{\partial L}{\partial \mathbf{H_{i}}} \label{eq:8} \\
		\frac{\partial L}{\partial \mathbf{H_{i-1}}}&=&\frac{\partial L}{\partial \mathbf{Z_{i-1}}}\frac{\partial \mathbf{Z_{i-1}}}{\partial \mathbf{H_{i-1}}}=\mathbf{A_i^T}\frac{\partial L}{\partial \mathbf{H_{i}}}\odot \sigma'({\mathbf{H_{i-1}}}) \label{eq:9}
	\end{eqnarray}
	
	where $\odot$ denotes the element-wise product. Based on these equations, we can derive the relationship between the gradients in adjacent layers:
	\begin{equation}
		\frac{\partial L}{\partial \mathbf{A_{i-1}}} = \mathbf{A_i^T}(\frac{\partial L}{\partial \mathbf{A_{i}}}{(\mathbf{Z_{i-1}^T)^{-1}}}\odot \sigma'({\mathbf{H_{i-1}}})){\mathbf{Z_{i-2}}^T} \label{eq:rec}
	\end{equation}
	Define : $\mathbf{\epsilon_i} = \frac{\partial L}{\partial \mathbf{A_{i}}} - \bar\frac{\partial L}{\partial \mathbf{A_{i}}}$, where $\bar\frac{\partial L}{\partial \mathbf{A_{i}}}$ is the gradient calculated by the prototype $\bar X$. According to the definition for $\epsilon_i$, we have $\mathbb{E}[\mathbf{\epsilon_i}] = \mathbf{0}$ and the divergence for model in layer $i$ is $||\mathbf{\epsilon_i}||$. Rewrite Eq. \ref{eq:rec} using this definition, we have:
	\begin{equation}
		\frac{\partial L}{\partial \mathbf{A_{i-1}}} = \mathbf{A_i^T}(\epsilon_i{\mathbf{(Z_{i-1}^T)^{-1}}}\odot \sigma'({\mathbf{H_{i-1}}})){\mathbf{Z_{i-2}^T}} \label{eq:error_rec} + \mathbf{A_i^T}(\bar\frac{\partial L}{\partial \mathbf{A_{i}}}{\mathbf{(Z_{i-1}^T)^{-1}}}\odot \sigma'({\mathbf{H_{i-1}}})){\mathbf{Z_{i-2}^T}} \label{eq11}
	\end{equation}
	Now, we examine the phenomenon of divergence accumulation, which forms the basis of our theorem. To facilitate our analysis, we introduce two mild assumptions.
	
	\emph{Assumption 1:} The expectation of divergence is retained during the back-propagation process, that is: $\mathbb{E}[||\mathbf{A_i^T}(\mathbf{\epsilon_i}{(\mathbf{Z_{i-1}^T)^{-1}}}\odot \sigma'({\mathbf{H_{i-1}}})){\mathbf{Z_{i-2}}}^T||^2]$ = $\mathbb{E}[||\mathbf{\epsilon_i}||^2]$. 
	
	Assumption 1 can be readily met since various techniques are employed during the training of neural networks to maintain the gradient norm and mitigate issues such as gradient explosion or vanishing. These techniques include parameter initialization \cite{xavier, kaiming}, residual links \cite{resnet}, and normalization \cite{bn, ln}. Consequently, the back-propagated divergence can also be retained.
	
	\emph{Assumption 2:} The random variable vector $\mathbf{\epsilon_i}$ is independent with $\mathbf{Z}$ and $\mathbf{H}$.
	
	\emph{Theorem 1:} Given Assumption 1 and Assumption 2, it follows that the divergence of previous layer is no smaller than that of the later layer. Formally, this can be expressed as: $\mathbb{E}[||\mathbf{\epsilon_{i-1}}||^2)] \ge \mathbb{E}[|\mathbf{\epsilon_i}||^2]$. 
	
	The proof is outlined as follows: the expected divergence of layer $i-1$ is comprised of two components. The first is the divergence back-propagated from layer $i$ and the second is the divergence arising from distinct inputs $\mathbf{Z}$ and $\mathbf{H}$. This structure guarantees that divergence will consistently increase during back-propagation. A detailed proof is available in the appendix.
	
	The above theorem connects data diversity and model divergence in FL. Through the layer-by-layer back-propagation process, the divergences of subsequent layers accumulate on top of the divergences of previous layers. This accumulation is more pronounced in deeper networks due to their longer chains of divergence. Consequently, deeper networks tend to exhibit higher divergences, resulting in a decline in performance. This analysis provides an explanation for the observations presented in Section \ref{introduction}. Let us recapitulate the observations along with their corresponding explanations.
	
	\textbf{Observation 1:} Deeper models are challenging to converge.
	\textbf{Explanation:} In deeper models, gradients need to be calculated through a long chain of back-propagation. The divergence accumulates at each step of back-propagation and leads to a significant divergence in the final gradient, ultimately causing a decline in the performance of the model.
	
	\textbf{Observation 2:} Shallow layers tend to converge, while deep layers tend to diverge.
	\textbf{Explanation:} Referring to Eq. \ref{eq:rec}, the divergence for layer $i$ is influenced by its input $\mathbf{Z_{i-2}}$. If the previous layers do not converge, the inputs $\mathbf{Z_{i-2}}$ in different clients are computed using different previous layers and different data, leading to significant divergences. However, when the previous layers do converge, indicating high similarity among their preceding layers, the inputs $\mathbf{Z_{i-2}}$ have lower divergences. This also explains our earlier observation that the latter layers do not converge before the former ones.
	
	\subsection{Empirical Validation of Divergence Accumulation}
	We present a series of experiments to validate the phenomenon of divergence accumulation discussed in the previous subsection.  Our aim is to demonstrate two key points: 1) The accumulation of divergence during the back-propagation process, and 2) The larger divergence in preceding layers of deeper models due to their longer accumulation chain. Two approaches are employed to reflect the diversity of data across different clients.
	
	Firstly, we utilize images with Gaussian noises to model data diversity. We begin by calculating the gradient using an original image. Then, we add random Gaussian noises (drawn from $\mathcal{N}$(0, 0.1)) to the input image and calculate the corresponding gradients. We sample the noises 1000 times, and the standard deviation among these gradients precisely represents the model divergence when updating the same model with different image data.
	
	Secondly, we use images from the same class to model data diversity. We individually calculate the gradients for all images belonging to the same class. Similarly, we use the standard deviation among these gradients to reflect the model divergence resulting from different image training data.
	
	In our experiments, we employ two types of models, namely shallow and deep models, for calculations. To ensure the robustness of the results, each experiment is repeated three times using different random seeds, and we provide the confidence interval in each figure.
	
\begin{figure*}[!t]
		\centering
		\subfloat[Logged divergences for data with Gaussian noises.]{\includegraphics[width=2.75in]{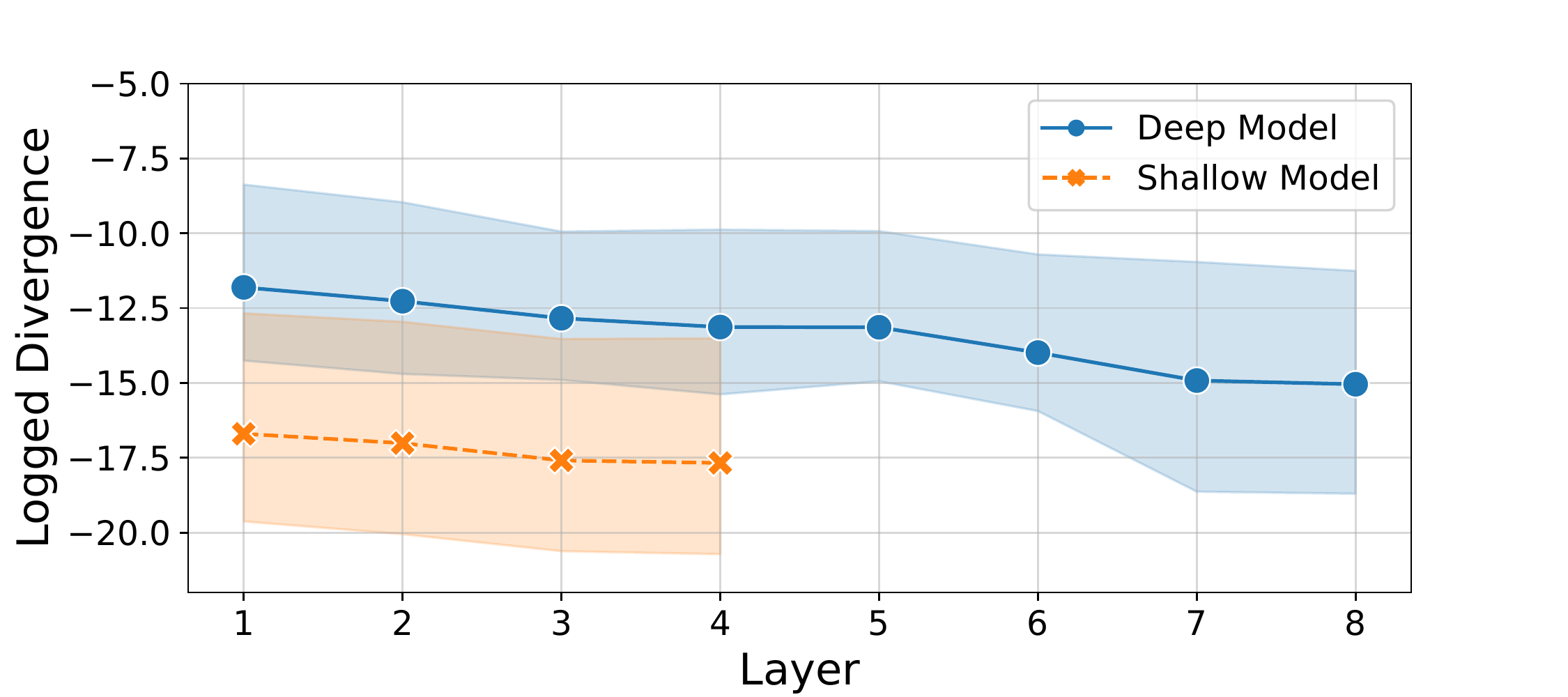}%
			\label{fig:backward_std_gauss}}
		\hfil
		\subfloat[Logged divergences for data in the same class.]{\includegraphics[width=2.75in]{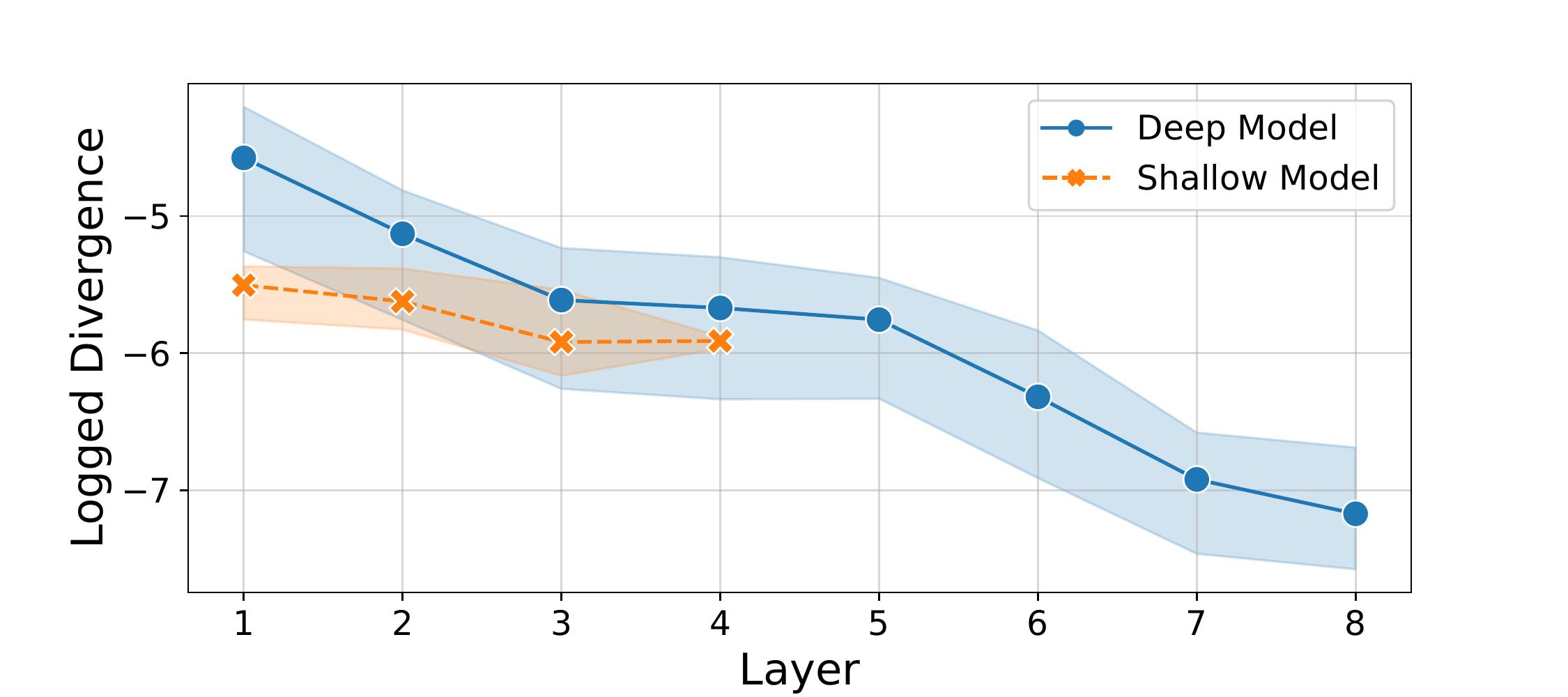}%
			\label{fig:backward_std_real}}
		\caption{In this experiment, we demonstrate the divergences of a 8-layer CNN (deep model) and a 4-layer CNN (shallow model) trained on MNIST. We use the log function $f(x) = \log x$ to transform divergences into the same scale. The dark-colored lines represent the mean of divergence within different classes, and the light-colored areas represent the range of real divergences.}
		\label{fig:diff_std_real_gauss}
	\end{figure*}

	Divergences under the two types of data diversities are plotted layer-wisely in Fig. \ref{fig:diff_std_real_gauss}. The plots clearly illustrate that, for both types of data diversity, the divergence within the same model accumulates during the back-propagation process. When comparing the shallow model and the deep model, it is evident that the shallow model exhibits lower divergence in preceding layers due to its shorter accumulation chain. The findings indicate that the phenomenon of divergence accumulation is a general conclusion that holds across different scenarios. 
	
	\section{Enhancing the Performance of Deeper Neural Networks in FL}
	\label{section:experiment}
	
	In this section, we put forth a set of guidelines to enhance the performance of FL when utilizing deeper neural networks. As the divergence of models plays a crucial role in determining the effectiveness of FL, the central goal is therefore to minimize these divergences during the FL training process.
	
	As shown in Eq. \ref{eq:error_rec}, the two terms on the right hand side of the equation represent the dual factors influencing the divergence for a specific layer: the divergence back-propagated from the subsequent layer and the divergence originating from distinct inputs. These factors contribute independently to the overall divergence. The first component, related to the model architecture, plays a significant role in influencing the divergence. We will demonstrate that specific architectural designs, such as utilizing wider models with smaller receptive fields, can effectively reduce model divergences. The second component is associated with the input data. We will illustrate that by increasing the similarity of data across different clients, we can correspondingly reduce the model divergences. Augmenting the similarities of data among clients in FL leads to a decrease in divergences, thereby significantly enhancing the performance of FL when utilizing deeper models.
	
	In the following subsections, we will introduce guidelines and present corresponding empirical evidence. First, we introduce the general experimental settings. We use the Tiny-ImageNet dataset\cite{tinyimagenet} (10,000 training images, 1,000 test images), the CIFAR100 dataset\cite{cifar100} (50,000 training images, 10,000 test images), and the CIFAR10 dataset\cite{cifar10} (50,000 training images, 10,000 test images). For the default data preprocessing, we resize all images to a size of 64$\times$64 and applied Random Resized Crop (RRC) as data augmentation on the training set. We set the number of clients to 30, the number of local training epochs to 8, and the learning rate to 0.02. We utilize FedAvg as the baseline algorithm as the data on each client are i.i.d.  To effectively evaluate the performance of various designs, we employ two key metrics: Test Accuracy (denoted as Acc.) and Mean Divergence (denoted as Div.).
	In terms of the construction of experimental platform, we simulate multiple clients with a single A100 GPU and conducted simulation tests on it. The code has been made public.

	\subsection{Guidelines for Enhancing Model Architectures}
	\label{subsection:model_arch}
	
	In this subsection, we propose several guidelines to decrease model divergence across various clients. As the accumulation of divergence is associated with back-propagation and consequently linked to the model's architecture, we can reduce divergence by strategically designing the model architecture. 
	
	\textbf{Using wider models.}
	In centralized learning, prior research has demonstrated the benefits of utilizing "wider" network models. In the context of FL, our experimental findings, as presented in Table 1, confirm that wider neural networks yield greater enhancements compared to centralized learning. Notably, we observe a reduction in divergence among clients when employing wider networks. This phenomenon can be attributed to the "lazy" characteristic exhibited by wider neural networks, as suggested by previous studies \cite{lazy-model}. The reduced parameter changes within each network contribute to a decrease in divergence among models across different clients. Consequently, this reduction in divergence fosters improved FL performance, aligning with our overarching objective.
	
	\begin{table}[t]
		\label{tab:width}
		\caption{Experiments conducted using a 13-layer ResNet with different widths. Acc. is the best model accuracy at test time and Div. is the mean divergence of the model of all layers.}
		\centering
		\begin{tabular}{c | c | cc | cc | cc}
			\toprule
			\multirow{2}{*}{Dataset} &
			\multirow{2}{*}{Method} &
			\multicolumn{2}{c|}{\textbf{1$\times$width}}  &
			\multicolumn{2}{c|}{\textbf{2$\times$width}}  &
			\multicolumn{2}{c}{\textbf{3$\times$width}} 
			\\
			&
			& Acc.&Div.
			& Acc.&Div.
			& Acc.&Div. \\
			\midrule
			\multirow{2}{*}{Tiny-Imagenet} &
			{CL} 
			&0.509 &-
			&0.526 &-
			&0.546 &-
			\\
			& {FL} 
			&0.458 & 0.022
			&0.484 & 0.012
			&0.504 & 0.006
			\\
			\midrule
			\multirow{2}{*}{CIFAR100} &
			{CL} 
			&0.587 &-
			&0.623 &-
			&0.634 &-
			\\
			& {FL} 
			&0.527 & 0.025
			&0.596 & 0.016
			&0.613 & 0.008
			\\
			\bottomrule
		\end{tabular}
	\end{table}
	
	\textbf{Using models with smaller receptive fields.} 
	The receptive field refers to the region of the input image that influences the activation of a particular neuron and is important in CNNs. Neurons with smaller receptive fields predominantly extract low-level semantic information, while those with larger receptive fields focus on high-level semantic information. The receptive field is calculated using the recursive formula $l_k=l_{k-1}+[(K_k - 1)\prod_{i=1}^{k-1}s_i]$, where $l_k$ represents the receptive field of the $k$-th layer, $K_k$ and $s_i$ denote the kernel size and stride of the $k$-th layer, respectively.
	
	To explore the impact of receptive field size on FL performance, we conducted experiments with two settings aimed at reducing the receptive fields. The results, presented in Table 2, reveal that minimizing the receptive field effectively limits the divergence between models, thereby enhancing the overall FL performance. This observation is attributed to the fact that a smaller receptive field ensures that neurons primarily observe similar image regions, such as low-level semantic information like edges and corners. By reducing the diversity of input data, we effectively reduce model divergence, as indicated by our earlier analysis.
	
	Additionally, we assessed the performance of two vision transformer models, ViT \cite{vit} and Swin \cite{swin}, which offer contrasting approaches to receptive field design. While ViT boasts infinitely large receptive fields for neurons in each layer, Swin maintains a specific receptive field size. We conducted experiments using both models, each comprising six transformer blocks and a parameter count similar to ResNet18. The experimental results, presented in Table 3, demonstrate that the Swin transformer exhibits a relatively smaller performance decline in FL compared to the ViT model. This finding further validates our guideline of utilizing models with small receptive fields.
	
	\begin{table}[t]
		\label{tab:recep}
		\caption{Experiments conducted using ResNet101 with different receptive fields. We tried two settings: replacing the first 7$\times$7 convolution with a 3$\times$3 one \textbf{(3$\times$3 Conv.)}, and removing the MaxPooling layer \textbf{(No M.P.)}. Both settings reduce the receptive field of subsequent convolutional layers. }
		\centering
		\begin{tabular}{c | c | cc | cc | cc }
			\toprule
			\multirow{2}{*}{Dataset} &
			\multirow{2}{*}{Method} &
			\multicolumn{2}{c|}{\textbf{Base}}  &
			\multicolumn{2}{c|}{\textbf{3$\times$3 Conv.}}  &
			\multicolumn{2}{c}{\textbf{No M.P.}} 
			\\
			&
			& Acc.&Div.
			& Acc.&Div.
			& Acc.&Div. \\
			\midrule
			\multirow{2}{*}{Tiny-Imagenet} &
			{CL} 
			&0.340 &-
			&0.472 &-
			&0.486 &-
			\\
			& {FL} 
			&0.236 & 0.005
			&0.457 & 0.004
			&0.473 & 0.004
			\\
			\midrule
			\multirow{2}{*}{CIFAR100} &
			{CL} 
			&0.510 &-
			&0.594 &-
			&0.602 &-
			\\
			& {FL} 
			&0.413 & 0.006
			&0.574 & 0.006
			&0.543 & 0.007
			\\
			\midrule
			\multirow{2}{*}{CIFAR10} &
			{CL} 
			&0.728 &-
			&0.775 &-
			&0.793 &-
			\\
			& {FL} 
			&0.618 & 0.007
			&0.701 & 0.006
			&0.724 & 0.005
			\\
			\bottomrule
		\end{tabular}
	\end{table}
	
	\begin{table}[t]
		\label{tab:vit}
		\caption{Experiments conducted using ViT and Swin with 6 transformer blocks. For ViT, we choose patch size to be 4 with dimension 512. For Swin the patch size is 4 with window size 4. The depth and parameter number of these models are similar and compatible to ResNet18. Numbers in brackets refer to the ratio of performance degradation.}
		\centering
		\begin{tabular}{c | cc | cc | cc }
			\toprule
			\multirow{2}{*}{Model} &
			\multicolumn{2}{c|}{\textbf{Tiny-Imagenet}}  &
			\multicolumn{2}{c|}{\textbf{CIFAR100}}  &
			\multicolumn{2}{c}{\textbf{CIFAR10}} 
			\\
			
			& CL & FL
			& CL & FL
			& CL & FL \\
			\midrule
			{ViT}
			&0.340 &0.236 ($\downarrow$30.5\%)
			&0.572 &0.454 ($\downarrow$20.6\%)
			&0.684 &0.572 ($\downarrow$15.8\%)
			\\
			\midrule
			{Swin}
			&0.389 &0.326 ($\downarrow$16.2\%)
			&0.499 &0.433 ($\downarrow$13.2\%)
			&0.663 &0.607 ($\downarrow$8.45\%)
			\\
			\bottomrule
		\end{tabular}
	\end{table}
	
	\subsection{Guidelines for Optimizing Data Pre-processing}
	
	In this subsection, we delve into guidelines focused on data-centric approaches. The primary driver of model divergence stems from the diversity of data sourced from various clients. By implementing suitable pre-processing strategies for data, this divergence can be greatly decreased.
	
	\textbf{Using image with higher resolution if possible.} 
	Another effective approach to reducing model divergence is by utilizing images with higher resolutions. Interestingly, adjusting the image resolution and modifying the model's receptive field are two distinct implementation methods that share a common underlying principle. While reducing the receptive field allows individual neurons to focus on smaller pixel areas, increasing the image resolution results in less image information contained within the same-sized pixel area. Both methods fundamentally serve the same purpose. The results presented in Table 4 confirm our expectations, demonstrating that FL models perform better when operating on higher-resolution images. 
	
	\begin{table}[t]
		\label{tab:resolution}
		\caption{Experiments conducted using ResNet101 with different image resolution. We select the image sizes to be \textbf{64$\times$64}, \textbf{128$\times$128}, \textbf{196$\times$196}, respectively. }
		\centering
		\begin{tabular}{c | c | cc | cc | cc}
			\toprule
			\multirow{2}{*}{Dataset} &
			\multirow{2}{*}{Method} &
			\multicolumn{2}{c|}{\textbf{64$\times$64}}  &
			\multicolumn{2}{c|}{\textbf{128$\times$128}}  &
			\multicolumn{2}{c}{\textbf{192$\times$192}} 
			\\
			&
			& Acc.&Div.
			& Acc.&Div.
			& Acc.&Div. \\
			\midrule
			\multirow{2}{*}{Tiny-Imagenet} &
			{CL} 
			&0.340 &-
			&0.495 &-
			&0.540 &-
			\\
			& {FL} 
			&0.236 & 0.005
			&0.460 & 0.004
			&0.490 & 0.003
			\\
			\midrule
			\multirow{2}{*}{CIFAR100} &
			{CL} 
			&0.510 &-
			&0.604 &-
			&0.645 &-
			\\
			& {FL} 
			&0.413 & 0.006
			&0.501 & 0.006
			&0.603 & 0.005
			\\
			\midrule
			\multirow{2}{*}{CIFAR10} &
			{CL} 
			&0.728 &-
			&0.785 &-
			&0.799 &-
			\\
			& {FL} 
			&0.618 & 0.007
			&0.697 & 0.006
			&0.754 & 0.005
			\\
			\bottomrule
		\end{tabular}
	\end{table}
	
	\textbf{Using proper data augmentation methods.}
	By adopting appropriate data augmentation methods, it is possible to enhance the similarity of data within each client, leading to a reduction in model divergences. In our experiments, we employed two popular techniques: Random-Resized-Crop (RRC) and Color-Jitter (CJ). RRC involves randomly cropping a region from an image and subsequently resizing it back to its original dimensions. On the other hand, CJ refers to randomly altering the color attributes of the image. The experimental results, as presented in Table 5, reveal the significant improvements achieved by appropriate data augmentation techniques, particularly RRC, in enhancing the performance of FL models. However, it is crucial to strike a balance when applying data augmentation. Excessive or inappropriate augmentation approaches can lead to a slowdown in local training speed and result in a degradation of model accuracy. Therefore, it is important to carefully select and calibrate the data augmentation methods to achieve optimal performance in FL models.
	
	\begin{table}[h]
		\label{tab:data_aug}
		\caption{Experiments on ResNet101 with 4 augmentation methods: no augmentation  \textbf{(None)}, pure Random Resized Crop \textbf{(RRC)}, pure Color Jitter \textbf{(CJ)} and mixture of RRC and CJ \textbf{(Both)}. }
		\centering
		\begin{tabular}{c | c | cc | cc | cc | cc}
			\toprule
			\multirow{2}{*}{Dataset} &
			\multirow{2}{*}{Method} &
			\multicolumn{2}{c|}{\textbf{None}}  &
			\multicolumn{2}{c|}{\textbf{RRC}}  &
			\multicolumn{2}{c|}{\textbf{CJ}} &
			\multicolumn{2}{c}{\textbf{Both}} 
			\\
			&
			& Acc.&Div.
			& Acc.&Div.
			& Acc.&Div.
			& Acc.&Div. \\
			\midrule
			{Tiny} &
			{CL} 
			&0.286 &-
			&0.340 &-
			&0.218 &-
			&0.321 &-
			\\
			{ImageNet}
			& {FL} 
			&0.224 & 0.012
			&0.236 & 0.005
			&0.147 & 0.006
			&0.252 & 0.006
			\\
			\midrule
			\multirow{2}{*}{CIFAR100} &
			{CL} 
			&0.393 &-
			&0.510 &-
			&0.330 &-
			&0.455 &-
			\\
			& {FL} 
			&0.269 & 0.011
			&0.413 & 0.006
			&0.235 & 0.008
			&0.350 & 0.007
			\\
			\bottomrule
		\end{tabular}
	\end{table}
	
	\section{Conclusion and Future Work}
	In this paper, we observe that deeper neural networks are difficult to converge in FL, which we believe is a critical problem for large-scale FL. To gain a deeper understanding of this problem and propose a solution, we introduce and examine the concept of divergence accumulation. Finally, several guidelines are proposed to reduce the divergence, which greatly improved the performance of FL on deeper models. We believe that this work holds significant value and serves as a source of inspiration for future research in large-scale FL. Building upon the theoretical foundations established in this paper, our future endeavors will involve developing and presenting additional techniques that further improve the performance of deeper networks within the FL paradigm.
	
	\bibliographystyle{elsarticle-num}
	\bibliography{mylib}

\newpage
\section*{Appendix}
\section*{A. Proof for Divergence Accumulation}
In this section, we provide a detailed proof of \emph{Theorem 1} proposed in Section \ref{theorem}. Let us review the content and relevant assumptions of the theorem first:

\emph{Assumption 1:} The expectation of divergence is retained during the back-propagation process, that is:
\begin{equation}
	\mathbb{E}[||\mathbf{A_i^T}(\mathbf{\epsilon_i}{(\mathbf{Z_{i-1}^T)^{-1}}}\odot \sigma'({\mathbf{H_{i-1}}})){\mathbf{Z_{i-2}}}^T||^2] = \mathbb{E}[||\mathbf{\epsilon_i}||^2]
\end{equation}

\emph{Assumption 2:} The random variable matrix $\mathbf{\epsilon_i}$ is independent with $\mathbf{Z}$ and $\mathbf{H}$.

\emph{Theorem 1:} Given Assumption 1 and Assumption 2, it follows that the divergence of previous layer is no smaller than that of the later layer. Formally, this can be expressed as:
\begin{equation}
	\mathbb{E}[||\mathbf{\epsilon_{i-1}}||^2)] \ge \mathbb{E}[|\mathbf{\epsilon_i}||^2]
\end{equation}

\emph{Proof:} According to the definition of $\mathbf{\epsilon_{i-1}}$, we have:
\begin{equation}
||\mathbf{\epsilon_{i-1}}||^2 = ||\frac{\partial L}{\partial \mathbf{A_{i-1}}} - \frac{\bar {\partial L}}{\partial \mathbf{A_{i-1}}}||^2
\end{equation}

Using Eq. \ref{eq11} to rewrite this equation:
\begin{eqnarray}
||\mathbf{\epsilon_{i-1}}||^2 = ||\mathbf{A_i^T}(\epsilon_i{\mathbf{Z_{i-1}^{-1}}}\odot \sigma'({\mathbf{H_{i-1}}})){\mathbf{Z_{i-2}}} + \mathbf{A_i^T}(\bar\frac{\partial L}{\partial \mathbf{A_{i}}}{\mathbf{Z_{i-1}^{-1}}}\odot \sigma'({\mathbf{H_{i-1}}})){\mathbf{Z_{i-2}}} - \frac{\bar {\partial L}}{\partial \mathbf{A_{i-1}}}||^2 
\end{eqnarray}

For simplicity, we use $\mathcal{T}_1$ and $\mathcal{T}_2$ to represent the two terms on the right hand side of the equation above, and flatten them into column vectors, that is:
\begin{eqnarray}
	\mathcal{T}_1 &=& Flatten(\mathbf{A_i^T}(\epsilon_i{\mathbf{Z_{i-1}^{-1}}}\odot \sigma'({\mathbf{H_{i-1}}})){\mathbf{Z_{i-2}}}) \\
	\mathcal{T}_2 &=& Flatten(\mathbf{A_i^T}(\bar\frac{\partial L}{\partial \mathbf{A_{i}}}{\mathbf{Z_{i-1}^{-1}}}\odot \sigma'({\mathbf{H_{i-1}}})){\mathbf{Z_{i-2}}} - \frac{\bar {\partial L}}{\partial \mathbf{A_{i-1}}})
\end{eqnarray}

So, we have:
\begin{equation}
	||\mathbf{\epsilon_{i-1}}||^2 = ||\mathcal{T}_1 + \mathcal{T}_2||^2 = ||\mathcal{T}_1||^2 + ||\mathcal{T}_2||^2 + \mathcal{T}_1^T\mathcal{T}_2+\mathcal{T}_2^T\mathcal{T}_1 \\
\end{equation}

By taking the expectation of both sides of the equation, we can derive:
\begin{equation}
\mathbb{E}[||\mathbf{\epsilon_{i-1}}||^2] = \mathbb{E}[||\mathcal{T}_1||^2] + \mathbb{E}[||\mathcal{T}_2||^2] + \mathbb{E}[\mathcal{T}_1^T\mathcal{T}_2]+\mathbb{E}[\mathcal{T}_2^T\mathcal{T}_1]
\end{equation}

Next, we will prove that $\mathbb{E}[\mathcal{T}_1^T\mathcal{T}_2] = 0$. Assume that $\mathbf{\epsilon_{i}} = (\epsilon_{i,j})\in \mathbb{R}^{d_1\times d_2}$, where $\mathbb{E}[\epsilon_{i,j}] = 0, \forall i \in [1, d_1], j \in [1, d_2]$. We first put forth some lemmas.

\emph{Lemma 1:} If $\mathbf{A}=(a_{i,j})\in\mathbb{R}^{d_1\times d_2}$, and each $a_{i,j}$ is a linear combination of $\epsilon_{p,q}$, that is: $a_{i,j} = \sum_{p,q} \alpha_{p,q}^{i,j}\epsilon_{p,q}$; $\mathbf{B}=(b_{i,j})\in\mathbb{R}^{d_2\times d_3}$, and each $b_{i,j}$ is independent of $\epsilon_{p,q}$;  $\mathbf{C}=(c_{i,j})\in\mathbb{R}^{d_4\times d_1}$, and each $c_{i,j}$ is independent of $\epsilon_{p,q}$. Then, each element in $\mathbf{C}\cdot\mathbf{A}$ and $\mathbf{A}\cdot\mathbf{B}$ is also a linear combination of $\epsilon_{p,q}$.

\emph{Proof:} For each element $d_{i,j}$ in $\mathbf{C}\cdot\mathbf{A}\in\mathbb{R}^{d_4\times d_2}$, we have
\begin{eqnarray}
	d_{i,j} &=& \sum_k c_{i,k}a_{k,j} = \sum_k c_{i,k} \sum_{p,q} \alpha_{p,q}^{k,j}\epsilon_{p,q} \\
	&=& \sum_k \sum_{p,q} c_{i,k}\alpha_{p,q}^{i,k}\epsilon_{p,q} = \sum_{p,q} \sum_{k} c_{i,k}\alpha_{p,q}^{k,j}\epsilon_{p,q} \\
	&=& \sum_{p,q} (\sum_{k} c_{i,k}\alpha_{p,q}^{k,j})  \epsilon_{p,q}
\end{eqnarray}

Hence, each element in $\mathbf{C}\cdot\mathbf{A}$ can be represented as a linear combination of $\epsilon_{p,q}$. Similarly, for each element $e_{i,j}$ in $\mathbf{A}\cdot\mathbf{B}\in\mathbb{R}^{d_1\times d_3}$, we have:
\begin{eqnarray}
	e_{i,j} &=& \sum_k a_{i,k}b_{k,j} = \sum_k b_{k,j} \sum_{p,q} \alpha_{p,q}^{i,k}\epsilon_{p,q} \\
	&=& \sum_k \sum_{p,q} b_{k,j}\alpha_{p,q}^{i,k}\epsilon_{p,q} = \sum_{p,q} \sum_{k} b_{k,j}\alpha_{p,q}^{i,k}\epsilon_{p,q} \\
	&=& \sum_{p,q} (\sum_{k} b_{k,j}\alpha_{p,q}^{i,k})  \epsilon_{p,q}
\end{eqnarray}
This finishes the proof for Lemma 1.

\emph{Lemma 2:} If $\mathbf{A}=(a_{i,j})\in\mathbb{R}^{d_1\times d_2}$, and $a_{i,j} = \sum_{p,q} \alpha_{p,q}^{i,j}\epsilon_{p,q}$; $\mathbf{B}=(b_{i,j})\in\mathbb{R}^{d_1\times d_2}$, and each $b_{i,j}$ is independent of $\epsilon_{p,q}$. Then, each element in $\mathbf{A}\odot\mathbf{B}$ is also a linear combination of $\epsilon_{p,q}$.

\emph{Proof:} For each element $f_{i,j}$  in $\mathbf{A}\odot\mathbf{B}\in\mathbb{R}^{d_1\times d_2}$, we have:
\begin{eqnarray}
	f_{i,j} &=& a_{i,j}b_{i,j} = b_{i,j}\sum_{p,q} \alpha_{p,q}^{i,j}\epsilon_{p,q} =\sum_{p,q}  (b_{i,j}\alpha_{p,q}^{i,j})\epsilon_{p,q}
\end{eqnarray}

Based on Lemma 1 and Lemma 2 above, we can conclude that $\mathcal{T}_1^T\mathcal{T}_2$ can be expressed as a linear combination of $\epsilon_{p,q}$, that is:
\begin{equation}
\mathcal{T}_1^T\mathcal{T}_2 = \sum_{p,q} \mathcal{C}_{p,q} \epsilon_{p,q}
\end{equation}

where each $\mathcal{C}_{p,q}$ does not contain any term associated with $\epsilon_{p,q}$. According to Assumption 2, the random variable $\mathcal{C}_{p,q}$ is independent of $\epsilon_{p,q}$, so we can simplify its expectation form as:
\begin{equation}
\mathbb{E}[\mathcal{T}_1^T\mathcal{T}_2] = \mathbb{E}[\sum_{p,q} \mathcal{C}_{p,q} \epsilon_{p,q}] = \sum_{p,q} \mathbb{E}[\mathcal{C}_{p,q} \epsilon_{p,q}] = \sum_{p,q} \mathbb{E}[\mathcal{C}_{p,q}]\mathbb{E}[\epsilon_{p,q}] = \sum_{p,q} \mathbb{E}[\mathcal{C}_{p,q}]\cdot 0 = 0
\end{equation}

By the same reasoning, we can conclude that $\mathbb{E}[\mathcal{T}_2^T\mathcal{T}_1] = 0$. Finally, we have:
\begin{eqnarray}
	\mathbb{E}[||\mathbf{\epsilon_{i-1}}||^2] &=& \mathbb{E}[||\mathcal{T}_1||^2] + \mathbb{E}[||\mathcal{T}_2||^2] + \mathbb{E}[\mathcal{T}_1^T\mathcal{T}_2]+\mathbb{E}[\mathcal{T}_2^T\mathcal{T}_1]\\ &=& \mathbb{E}[||\mathcal{T}_1||^2] + \mathbb{E}[||\mathcal{T}_2||^2]
\end{eqnarray}

According to Assumption 1, $\mathbb{E}[||\mathcal{T}_1||^2] = \mathbb{E}[||\mathbf{\epsilon_i}||^2]$, so:
\begin{eqnarray}
	\mathbb{E}[||\mathbf{\epsilon_{i-1}}||^2]-\mathbb{E}[||\mathbf{\epsilon_{i}}||^2]&=&\mathbb{E}[||\mathcal{T}_1||^2] -\mathbb{E}[||\mathbf{\epsilon_{i}}||^2] + \mathbb{E}[||\mathcal{T}_2||^2]\\ &=& \mathbb{E}[||\mathcal{T}_2||^2] \ge 0
\end{eqnarray}

This finishes the proof of Theorem 1.

\section*{B. Additional Experiments}
\subsection*{Detailed Layer-wise Divergences}
In Section \ref{section:experiment}, we demonstrate the Mean Divergence (Div.) under different experimental settings, but without showing the differences among different layers. Here, we provide more detailed experimental results for divergences across different layers.

The experiment settings in the following figures are the same as those in Section \ref{section:experiment}. Each figure caption indicates the dataset used and the number of layer divergence plotted.

\begin{figure}[h]
	\centering
	\includegraphics[width = 1.1\textwidth]{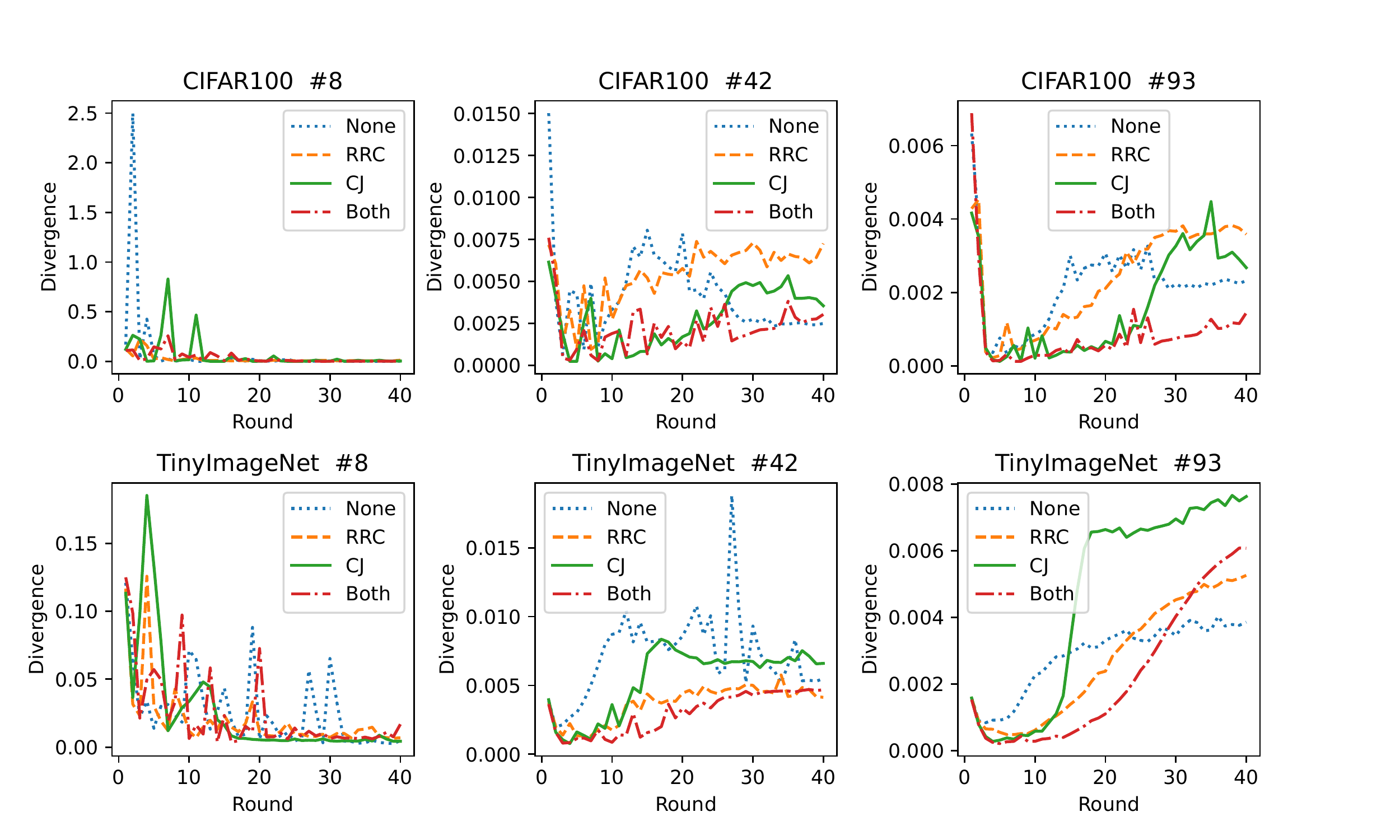}
	\caption{Divergences for different data-augmentation methods.}
	\label{fig:diff_aug}
\end{figure}

Fig. \ref{fig:diff_aug} illustrates divergences using different data-augmentation techniques. It is evident that by applying appropriate data-augmentation methods, we can obtain lower and more stable model divergences, ultimately enhancing performance.

\begin{figure}[t]
	\centering
	\includegraphics[width = 1.1\textwidth]{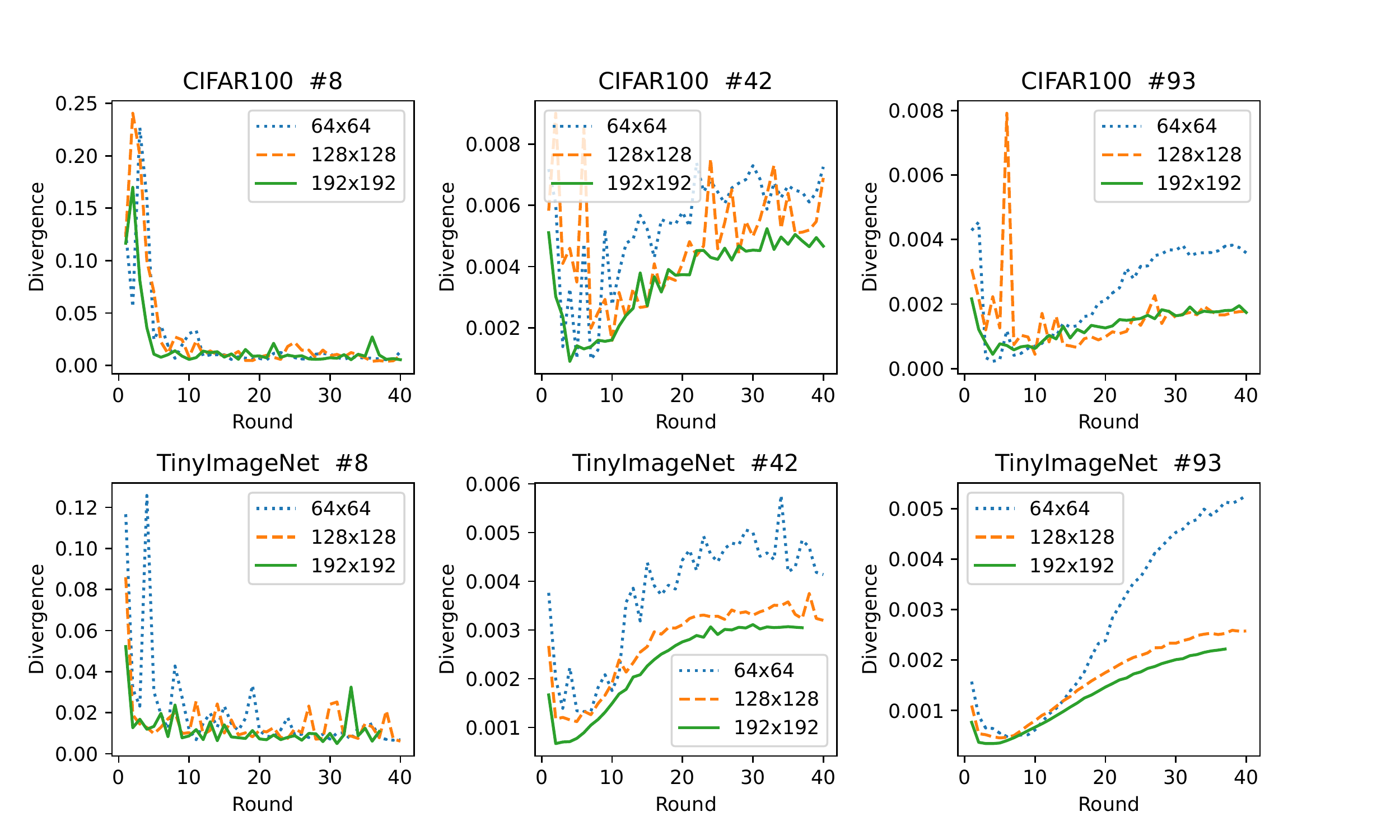}
	\caption{Divergences for different image resolutions.}
	\label{fig:diff_reso}
\end{figure}

Fig. \ref{fig:diff_reso} shows model divergences using different input image resolutions. We observe that when a smaller resolution is used, the model divergences for deep layers are significantly larger. This is because deeper layers tend to have larger receptive fields, and if the resolution is not sufficient, the receptive field of deep layers may cover the entire image, leading to greater data dissimilarity and larger divergences.

\begin{figure}[t]
	\centering
	\includegraphics[width = 1.1\textwidth]{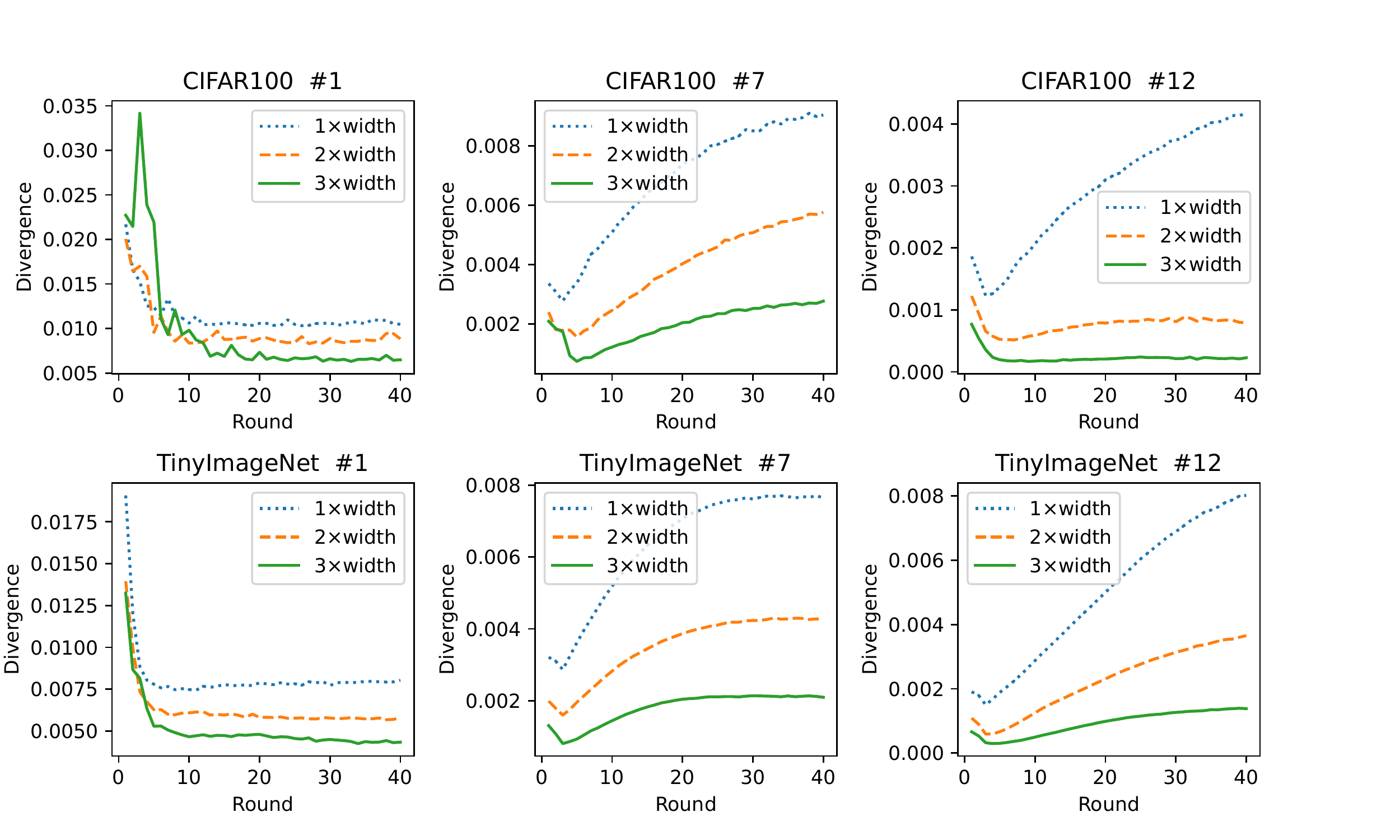}
	\caption{Divergences for different model width.}
	\label{fig:diff_width}
\end{figure}

Fig. \ref{fig:diff_width} presents model divergences using different model widths. It is clear that a wider model architecture results in a significant decline in divergence. This aligns with our expectation that wider models possess the "lazy" property, updating parameters more mildly.

\subsection*{Effect of Residual Link}
Residual link is a classic method in centralized learning, which can effectively improve the performance of deeper neural networks. In FL, as analyzed in Section \ref{theorem}, divergence accumulates and amplifies during the model’s back-propagation process. Residual connections can provide a "shortcut" for back-propagation, allowing gradients in deep layers to propagate quickly to shallower layers. As a result, since the length of the back-propagation path is reduced, the accumulation of model divergence is also reduced, and the convergence speed is faster. As can be seen from the experimental results in Table \ref{tab:res_link} although there is no significant difference in the final convergence accuracy of federated learning before and after adding residual connections, the convergence speed has increased.

	\begin{table}[h]
	\label{tab:res_link}
	\caption{We show different results with residual link and without residual link. Acc. is the best model accuracy at test time and \#Rnd. is the number of communication round when test accuracy first exceeds 0.4, 0.50. }
	\centering
	\begin{tabular}{c | c | cc | cc}
		\toprule
		\multirow{2}{*}{Dataset} &
		\multirow{2}{*}{Method} &
		\multicolumn{2}{c|}{\textbf{With Res. Link}}  &
		\multicolumn{2}{c}{\textbf{No Res. Link}}  
		\\
		&
		& Acc.&\#Rnd.
		& Acc.&\#Rnd. \\
		\midrule
		{Tiny} &
		{CL} 
		&0.509 &-
		&0.522 &-
		\\
		{ImageNet}
		& {FL} 
		&0.458 & 21
		&0.454 & 24
		\\
		\midrule
		\multirow{2}{*}{CIFAR100} &
		{CL} 
		&0.587 &-
		&0.569 &-
		\\
		& {FL} 
		&0.527 & 30
		&0.526 & 33
		\\
		\bottomrule
	\end{tabular}
\end{table}

\subsection*{Avoid Using Decreased Channel Dimensions}
In this part, we include an additional experiment. As mentioned in the introduction, we observe that the divergence of shallow layers tends to decrease and eventually converge, whereas the divergence of deep layers tends to intensify.

Intuitively, in order to leverage this property, we can assign a more significant role to the shallow layers compared to the deep layers. Conventionally, the number of channels in a CNN increases gradually as the layers go deeper. However, our proposed approach follows an opposite design principle where the number of channels gradually decreases.

For the experiment setup, we selected three models for comparison: a model with channel dimensions of (32, 64, 128, 256) following the typical design, a model with channel dimensions of (120, 120, 120, 120) referred to as the Mean design, and a model with channel dimensions of (256, 128, 64, 32) denoted as the Reversed design. In the subsequent results analysis, we refer to these designs as Normal, Mean, and Reversed, respectively.

\begin{figure}[h]
	\centering
	\includegraphics[width = 1.1\textwidth]{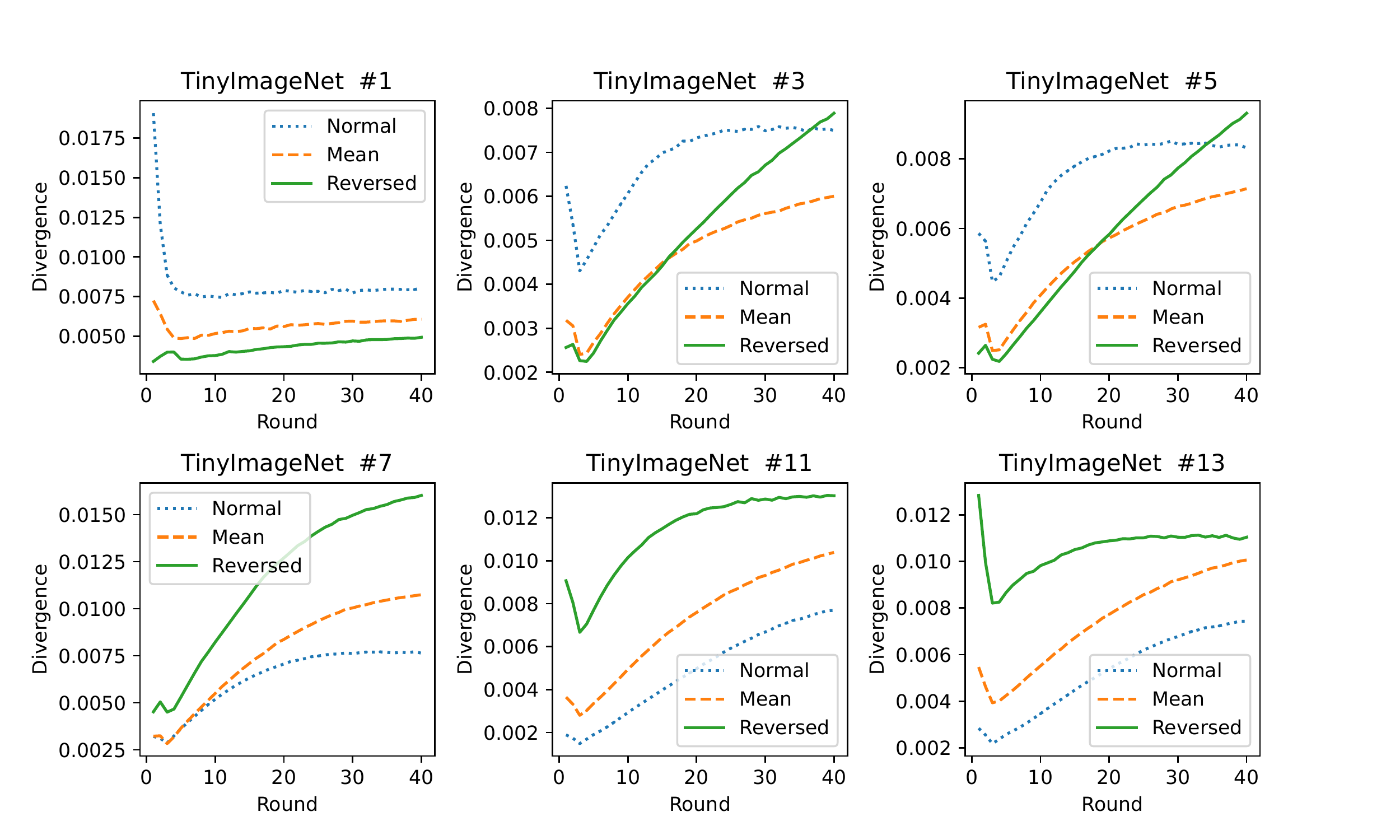}
	\caption{Divergences for different model architectures.}
	\label{fig:diff_seq}
\end{figure}

Surprisingly, we discovered that the Normal design, which adheres to the typical channel dimension pattern, exhibited the best performance with the lowest divergence. The final accuracies achieved by the three designs were 45.8\%, 44.0\%, and 35.4\%, respectively. To visualize the divergences across different layers, we plotted them layer-wise in Fig. \ref{fig:diff_seq}.

As depicted in the figure, the Reversed network displayed consistently larger divergences in the deep layers compared to the Normal network. Additionally, the divergences in the shallow layers of the Reversed network continuously intensified without showing any signs of convergence. This behavior can be attributed to the smaller width of the deeper layers, causing their divergences to be initially large due to the "lazy" property discussed in Section \ref{section:experiment}. Consequently, these large divergences accumulate onto the divergences of the shallow layers, resulting in a convergence issue.

\end{document}